\documentclass[10pt,twocolumn,letterpaper]{article}

\usepackage{iccv}
\usepackage{times}
\usepackage{epsfig}
\usepackage{graphicx}
\usepackage{amsmath}
\usepackage{amssymb}
\usepackage{multicol,multirow}
\usepackage{booktabs}
\usepackage{bbm}
\usepackage{caption}
\usepackage{subcaption}
\usepackage{colortbl}

\usepackage[breaklinks=true,bookmarks=false]{hyperref}

\iccvfinalcopy 

\def\iccvPaperID{****} 

\DeclareMathOperator*{\argmin}{argmin}
\begin{document}

\title{ConDA: Continual Unsupervised Domain Adaptation}
\makeatletter
\newcommand{\printfnsymbol}[1]{%
  \textsuperscript{\@fnsymbol{#1}}%
}
\makeatother
\author{Abu Md Niamul Taufique\thanks{equal contribution} \quad Chowdhury Sadman Jahan\printfnsymbol{1} \quad Andreas Savakis\\Rochester Institute of Technology \\ \{at7133,sj4654,andreas.savakis\}@rit.edu}




\maketitle

\begin{abstract}
Domain Adaptation (DA) techniques are important for overcoming the domain shift between the source domain used for training and the target domain where testing takes place.
However, current DA methods assume that the entire target domain is available during adaptation, which may not hold in practice. This paper considers a more realistic scenario, where target data become available in smaller batches and adaptation on the entire target domain is not feasible. 
In our work, we introduce a new, data-constrained DA paradigm where unlabeled target samples are received in batches and adaptation is performed continually. 
We propose a novel source-free method for continual unsupervised domain adaptation that utilizes a buffer for selective replay of previously seen samples.
In our continual DA framework, we selectively mix samples from incoming batches with data stored in a buffer using buffer management strategies and use the combination to incrementally update our model.
We evaluate the classification performance of the continual DA approach with state-of-the-art DA methods based on the entire target domain.
Our results on three popular DA datasets demonstrate that our method outperforms many existing state-of-the-art DA methods with access to the entire target domain during adaptation.

\end{abstract}

\section{Introduction}
\label{sec:introduction}

\begin{figure}[t]
\begin{minipage}[b]{1.0\linewidth}
  \centering
  \centerline{\includegraphics[width=\textwidth]{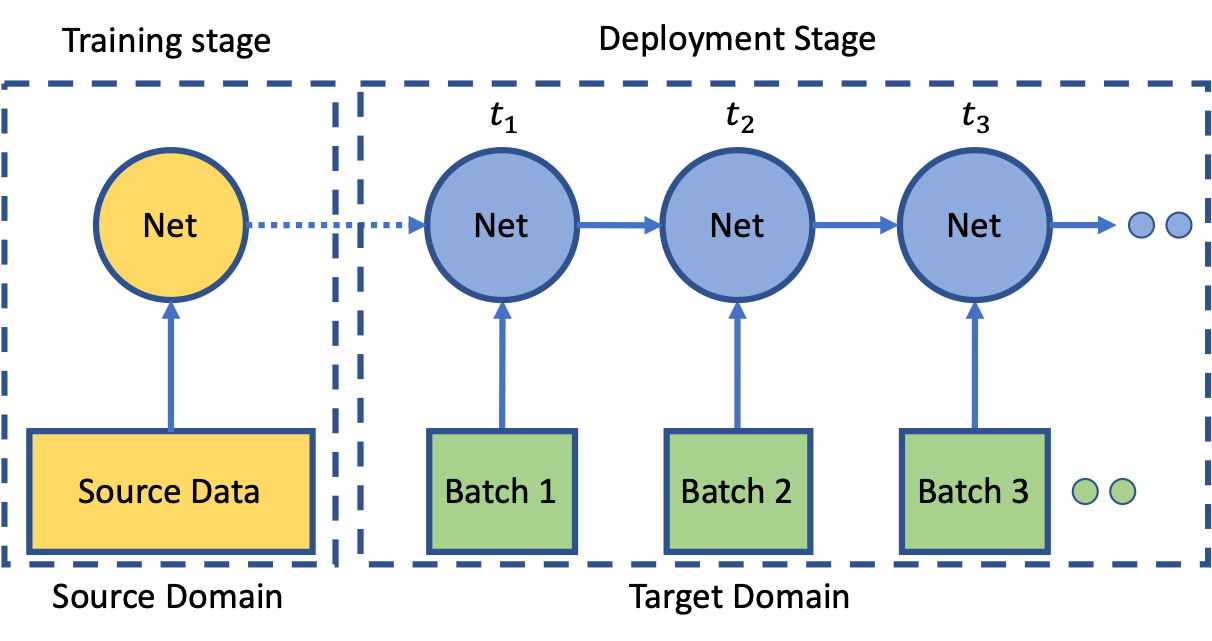}}
  \label{fig:process_v1}
\end{minipage}
\caption{Continual DA paradigm. Initial training is performed with labeled data in the source domain and the trained model is deployed in the target domain. During deployment, unlabelled target domain data are received in streaming batches and the model is continuously adapted with each new batch of target data.}
\label{fig:process}
\end{figure}


Domain adaptation (DA) methods based on deep learning have received significant attention in recent years for mitigating
the domain shift from the training domain (source) to the inference domain (target) \cite{ganin2015unsupervised,adda,gvb,shot,hdmi,mdd}.
In unsupervised domain adaptation (UDA), where the same classes are present in the source and target domains (closed set), the gap between the annotated source domain data and unlabeled target domain data is the main cause of reduction in classification accuracy. Many of the recent popular deep learning based DA methods \cite{gvb, alda, kurmi2019attending, cdan} employ adversarial training with both the source and target data to learn domain agnostic features, as proposed in \cite{ganin2015unsupervised}, or to align the feature spaces of the source and target domains, as was done in \cite{adda}. Inspired by Hypothesis Transfer Learning (HTL) \cite{kuzborskij2013stability}, some recent methods \cite{shot, kundu2020universal, hdmi} transfer only the source trained model for target adaptation, thus greatly reducing the data storage footprint. 

Current DA methods operate under the assumption that the entire target dataset is available during adaptation, which may not be feasible in practice, e.g., when an autonomous vehicle operates in a new environment. 
In this paper, we consider the scenario depicted in Figure \ref{fig:process}, where the network model is initially trained using source domain data and is then deployed in a new domain where target data are collected incrementally in small batches and the model is updated continually.  

In our continual DA framework, the shift between the source and target can be sudden and, depending on the datasets considered, the target distribution can be significantly different than the source distribution. 
In a related approach, Hoffman et al. \cite{hoffman2014continuous} proposed a manifold-based method that deals with streaming target data from an evolving target domain that is changing slowly. However, this work did not consider deep learning methods, it was not applied on standard DA datasets, and assumed there is no sudden domain shift between the source and target domains or between two consecutive time instances within the target domain. 
In contrast, we present a scenario where the target distribution is not directly related to the source distribution and the target data are received in a series of smaller batches over time, as shown in Figure \ref{fig:process}. Our approach is broader in scope, introduces a deep learning framework, and is applicable to standard DA datasets, making comparison with existing DA methods possible.

\begin{figure}
\centering
\begin{subfigure}[b]{0.45\textwidth}
   \includegraphics[width=0.95\linewidth]{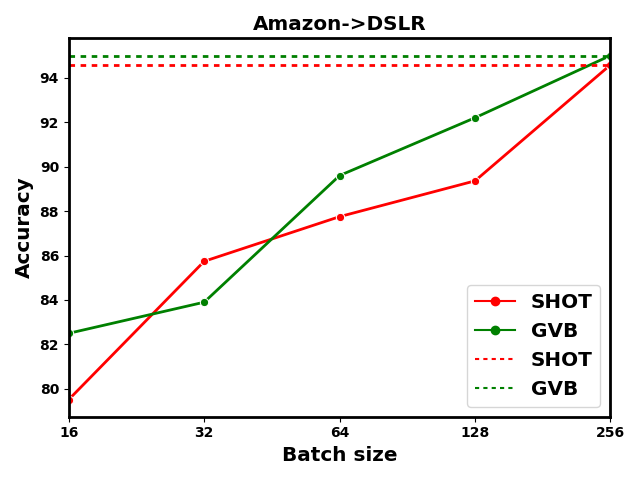}
   \label{fig:Ng1} 
\end{subfigure}
\begin{subfigure}[b]{0.45\textwidth}
   \includegraphics[width=0.95\linewidth]{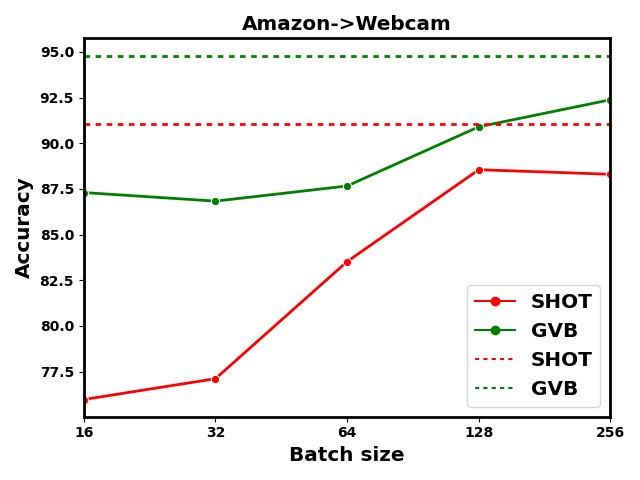}
   \label{fig:Ng2}
\end{subfigure}
\caption[]{Evaluation of SOTA DA methods SHOT \cite{shot} and GVB \cite{gvb} in a continual DA framework from Amazon to DSLR (top) and Amazon to Webcam (bottom). 
The horizontal dotted lines represent DA performance when the full target domain is available. 
The solid lines are based on adaptation using incoming target data batches of different size and performance evaluated on the entire target domain after all of the batches are seen by the network.
}
\label{fig:contda_results}
\end{figure}

To illustrate the impact of continual DA on current methods, we performed experiments in a continual setting with two state-of-the-art (SOTA) methods: Source Hypothesis Transfer (SHOT) \cite{shot}, a source-free DA method, and Gradually Vanishing Bridge (GVB) \cite{gvb}, a source dependent adversarial DA method. 
Using the Office-31 \cite{office} dataset, we considered Amazon as the source domain and adapted contiually on small batches of independent and identically distributed (i.i.d.) samples drawn from the target domain (DSLR or Webcam). 
The results of Figure \ref{fig:contda_results} indicate that continual adaptation performance drops significantly for both methods, compared to standard DA using the entire target data.  The drop in performance is more pronounced as the batch size gets smaller.

These results demonstrate the need for new continual DA approaches
that can maintain high performance while adapting to new target batches.
To solve this problem, we take cues from continual learning methods \cite{rebuffi2017icarl,wu2019large, hayes2020remind} and propose a Continual DA (ConDA) framework with a buffer to store processed samples and their predicted labels, and buffer management strategies to selectively store and replay previously seen target samples. Furthermore, our method incorporates better features with higher generalization capabilities that improve upon the performance of SOTA source-free DA methods. 

The ConDA approach continually adapts the source model to the target domain as data arrive in batches, which greatly reduces the data storage requirements. Our method does not require any source data during adaptation, and additionally does not need to store the whole target domain at any time. 
During adaptation, ConDA only requires the incoming batch of target data along with the data stored in the buffer. 
We evaluate several buffer configurations, along with specific loss functions for continual adaptation, and propose a buffer management strategy and associated adaptation procedure that is well-suited for continual DA. 
ConDA outperforms many non-continual DA methods that utilize the full target domain, yet it operates at a fraction of their data storage footprint. 
The main contributions of our paper are outlined below.

\begin{enumerate}
    \vspace{-0.05in}
    \item We propose a new paradigm for continual unsupervised domain adaptation performed on new batches of target samples.
    \vspace{-0.01in}
    \item We propose the ConDA framework for source-free continual DA that adapts on incoming batches of unlabeled target data and utilizes a buffer for selective replay of previous samples. 
    \vspace{-0.1in}
    \item During ConDA adaptation, we utilize sample mixup and equal diversity loss along with our buffer management strategy for effective adaptation. 
    \vspace{-0.1in}
    \item The performance of ConDA is superior to many SOTA DA methods, while utilizing a much smaller data storage footprint, even though other methods have access to the entire target and source domains. 
    \vspace{-0.1in}
    \item We demonstrate that high resolution features are useful for generalization across domains and achieve significant performance gains for UDA on standard datasets.
\end{enumerate}


\section{Related Work}
\label{sec:relatedwork}

\subsection{Unsupervised Domain Adaptation}
A domain gap manifests due to the dataset bias when the data distributions in the source and target domains are significantly different \cite{torralba2011unbiased}.
Many UDA techniques have been proposed to mitigate this domain gap for computer vision tasks such as object detection and semantic segmentation \cite{shot, khodabandeh2019robust, chen2019domain}. Long et al. \cite{long2015learning} and Tzeng et al. \cite{tzeng2014deep} proposed minimizing the maximum mean discrepancy (MMD) for UDA.
Zellinger et al. \cite{zellinger} proposed minimizing central moment discrepancy (CMD) by matching higher order central moments of probability distributions in the source and target data. Ganin et al. \cite{ganin2016domain} aligned distributions of source and target domains via an adversarial domain discriminator.
Tzeng et al. \cite{tzeng2017adversarial} adversarially aligned features of source and target domain data while transferring the source domain classifier to the target domain. Likewise, generative models have also been employed to create source-like images at the pixel level for domain adaptation \cite{cyclegan}. 

Adversarial methods require access to source data at the time of adaptation, but this is likely to create issues related to storage requirements or privacy when sharing of sensitive and private data. Domain adaptation research has been exploring such practical scenarios where adaptation is done without using source data. Source-free UDA methods consist of an initialization stage with access to source data for training and an adaptation stage with access only to the target data  without any of the source data \cite{kundu2020towards}. Chidlovskii et al. \cite{chidlovskii2016domain} proposed a semi-supervised source-free DA framework where no source domain data are available during adaptation, but some representation of the source domain is available, such as class means or a few annotated target samples. Liang et al. \cite{liang2019distant} identified a subspace where target and source centroids are only modestly shifted and used class-wise distribution estimator of the source data to conduct distant supervision for target adaptation. An end-to-end, source-free DA method based on information maximization was proposed in \cite{shot}.

\subsection{Continual Learning}

\begin{figure*}[htbp]
\begin{minipage}[b]{1.0\linewidth}
  \centering
  \centerline{\includegraphics[width=\textwidth]{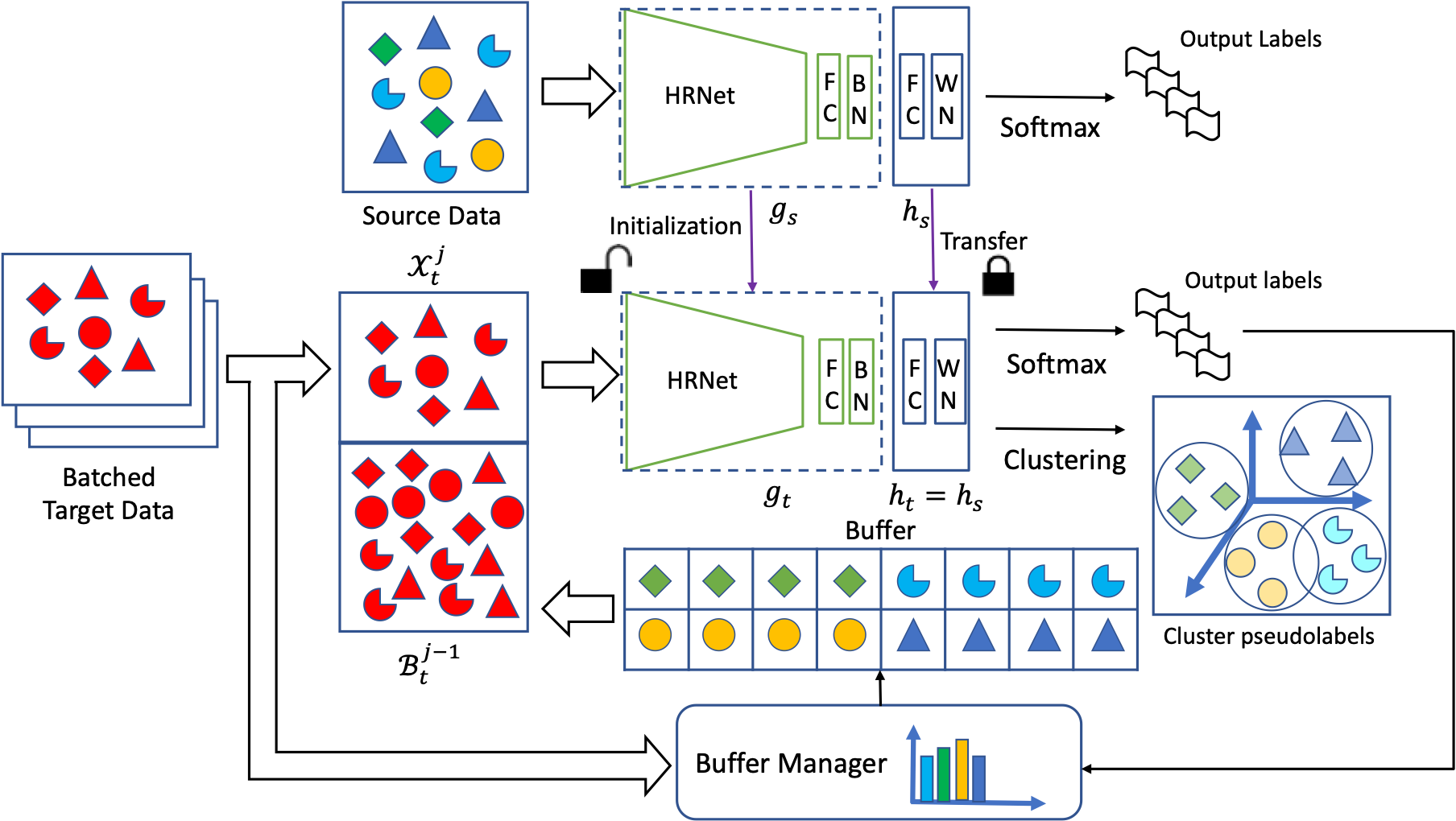}}
\end{minipage}
\caption{Proposed ConDA framework adapting on target domain data that arrive in small batches. A subset of the samples that are already seen by the network are stored in a buffer for replay with the incoming batches. The buffer manager is responsible for selecting the samples that populate the buffer. The incoming target samples are mixed with the current buffer samples and sent to the network for adaptation.}
\label{fig:architecture}
\end{figure*}

Mammals, as opposed to artificial neural networks trained within the standard deep leaning framework, learn continuously so that their intelligence increases gradually over time. When neural networks are subjected to such continual learning, they run the risk of catastrophic forgetting, where they forget the knowledge gained in earlier training stages \cite{mccloskey1989catastrophic}. Continual or lifelong learning methods have proposed a few mechanisms to mitigate catastrophic forgetting in deep neural networks. Among them, the most prominent are (i) replay of previously seen data \cite{rebuffi2017icarl, wu2019large, hayes2020remind}, (ii) constraining network parameter updates according to a regularization scheme \cite{kirkpatrick2017overcoming, zenke2017continual}, and (iii) network expansion with increasing data \cite{progressive, dynamic, distillation}. Memory replay mimics the mechanism of the human brain, where during both the sleeping \cite{sleep} and awake \cite{awake} phases, past experiences are regenerated from encoded representations and the neocortex is trained on them \cite{stickgold2001sleep,play}. Rebuffi et al. first applied memory replay in iCaRL  \cite{rebuffi2017icarl},  for class-incremental learning in the context of neural networks, where 20 raw samples from each class were stored for later replay. More recent replay methods extended iCaRL to make it end-to-end trainable \cite{endtoend}, introduced a loss function to correct for class bias \cite{wu2019large}, and stored mid-level features instead of raw images to reduce storage footprint \cite{hayes2020remind}. 
Regularization based models learn new tasks incrementally while preserving knowledge from previous tasks by varying the plasticity of the network's convolutional filter weights, which are significant for retaining earlier knowledge. 
Kirkpatric et al. \cite{kirkpatrick2017overcoming} proposed to selectively lower the learning rate from one task to the next. 

In this work, we mainly draw from the concept of memory replay. We present a way to continually adapt a source trained model to a new target domain when the target data are received in batches and not all available at the same time.  This is an area of domain adaptation that, to the best of our knowledge, has not yet explored. We showed in Figure \ref{fig:contda_results} that when incoming target data are received in batches, state-of-the-art DA methods suffer from performance degradation. We next present our ConDA method to overcome these limitations, and discuss strategies to configure the buffer and corresponding loss functions for continual DA. We benchmark our approach against standard DA methods and obtain SOTA results on some popular DA datasets. 


\section{Method}
\label{sec:method}
We consider a source domain $\mathcal{D}_s$ with labelled source samples $\{x_s^i, y_s^i\}_{i=1}^{n_s}$ where $n_s$ is the total number of
source samples $x_s^i \in \mathcal{X}_s$ with corresponding labels $y_s^i \in \mathcal{Y}_s$. We are given an unlabelled target domain $\mathcal{D}_t$ with $n_t$ samples $\{x_t^i\}_{i=1}^{n_t}$ and $x_t \in \mathcal{X}_t$. In closed-set UDA, we assume that the number of classes $\mathcal{C}_s$ present in the source domain is same as the number of classes $\mathcal{C}_t$ present in the target domain, and the task is to predict the target labels $\{y_t^i\}_{i=1}^{n_t}$ where $y_t \in \mathcal{Y}_t$. 
In the continual UDA setting, the target domain $\mathcal{D}_t$ is divided into $m$ batches, i.e., $\mathcal{X}_t = \{\mathcal{X}_t^1, \mathcal{X}_t^2, \mathcal{X}_t^3, .... , \mathcal{X}_t^m \}$   
with samples $\{x_t^{j,i}\}_{j=1, i=1}^{m, n_t^j}$ where $n_t^j$ is the number of samples in the $j^{th}$ batch and $j \in \{1, 2, 3, ....., m\}$.  
We consider that the source trained model $f_s: \mathcal{X}_s \rightarrow \mathcal{Y}_s$ is available with only a batch of target samples $\mathcal{X}_t^j$ at a time and our objective is to learn a model $f_t: \mathcal{X}_t^j \rightarrow \mathcal{Y}_t^j$ where $\mathcal{Y}_t^j$ is the predicted labels of $\mathcal{X}_t^j$. 

The continual DA scenario runs the risk of the model overfitting to the current batch of target samples and failing to adapt to the marginal distribution of the entire target domain due to the continual nature of the incoming samples. Therefore, our task is to reduce the performance gap between the model that is adapted based on continuous batches of target data, i.e., $f_t: \mathcal{X}_t^m \rightarrow \mathcal{Y}_t^m$ and the model that is adapted given the entire target domain simultaneously (standard DA framework), i.e., $f_t: \mathcal{X}_t \rightarrow \mathcal{Y}_t$, both evaluated on the full target domain $\mathcal{X}_t$.


Our ConDA framework for continual adaptation is shown in Figure \ref{fig:architecture}. The source model $f_s(x) = h_s(g_s(x))$ consists of two parts: a feature generator model $g_s$ that includes a backbone and a fully-connected layer followed by a batch normalization layer, and a hypothesis model $h_s$ that includes a fully connected layer and a weight normalization layer.
Inspired by \cite{shot}, we train the source model $f_s$
in a supervised manner with label smoothing \cite{muller2019does}. 
During target adaptation, we initialize the target hypothesis model with the source hypothesis, $h_t = h_s$, and the parameters of the hypothesis model remain unchanged over the adaptation procedure. We initialize the target feature generation model $g_t$ with the source feature generation model $g_s$ and adapt it with an incoming batch of target samples. 

In Section \ref{sec:introduction}, Figure \ref{fig:contda_results}, we showed that SOTA methods do not reach their full performance during continual adaptation. In ConDA, we propose to use a buffer to store selected target samples and replay them with the incoming batch of samples so that the network can generalize effectively over the entire target domain. 

\subsection{Buffer}
To conduct continual domain adaptation, we introduce a buffer $\mathcal{B}_t$ with states $\{\mathcal{B}_t^1, \mathcal{B}_t^2, ...., \mathcal{B}_t^m\}$ each corresponding to $m$ batches of target data. We maintain a class-balanced $\mathcal{B}_t$, i.e., an equal number of buffer slots are allocated for each class calculated from buffer length and the number of classes present in the target domain assuming that $\mathcal{C}_t = \mathcal{C}_s$. The buffer is populated after the network is trained on a batch of target samples. The buffer stores the samples and their corresponding class labels predicted by the network. Our model only requires access to the samples stored in the buffer for subsequent adaptation along with new target batches that arrive. The sample selection process to populate the buffer is handled by a buffer manager discussed in the following section.

\subsection{Buffer Manager}
Let's assume that the network is adapted on a batch $\mathcal{X}_t^j$ and outputs $\{\mathcal{Y}_t^j, \mathcal{U}_t^j\}$ where $\mathcal{U}_t$ is the softmax classification score. We compute the buffer sample labels $\mathcal{V}_t^{j-1}$ with the current state of the model $f_t: \mathcal{B}_t^{j-1} \rightarrow \mathcal{V}_t^{j-1}$. 
The buffer manager takes in $\{\mathcal{X}_t^j$,  $\mathcal{Y}_t^j$, $\mathcal{U}_t^j$, $\mathcal{B}_t^{j-1}$, and $\mathcal{V}_t^{j-1}\}$ and outputs $\mathcal{X}_t^\prime \subseteq \mathcal{X}_t^j \bigcup \mathcal{B}_t^{j-1}$ and corresponding labels to populate the buffer state $\mathcal{B}_t^j$. 
At first, the incoming batch samples are grouped based on the output label $\mathcal{Y}_t^j$, and samples of each class are sorted based on the confidence $\mathcal{U}_t^j$. Then, the buffer manager only picks the high confidence samples if the number of samples for any class exceeds the allotted number of slots for that class in the buffer. Finally, if available, the remaining space for that class is filled with randomly drawn samples from $\mathcal{B}_t^{j-1}$ of that class.

We conducted multiple experiments with a few other buffer selection techniques, such as choosing the incoming samples randomly, or selecting the buffer samples based on the cosine distance to the nearest self-supervised cluster centers. We did not find any significant performance variation with various buffer sample selection techniques. We found a slight increase in performance with the sample selection mechanism based on the higher confidence scores.

In the $(j+1)^{th}$ batch, the current buffer samples $\mathcal{B}_t^j$ and the incoming batch samples $\mathcal{X}_t^{j+1}$ are appended and provided to the network. 
We do not use any label information of the buffer samples when they are concatenated with the incoming batch samples.
During adaptation with the incoming batch and buffer samples, we performed clustering to compute pseudo labels. The clustering technique is described as follows.

\subsection{Clustering}
We adopted a self-supervised clustering method introduced in \cite{shot} as an extension of the Deep Cluster \cite{caron2018deep} method. The combination of the batch and the buffer samples is denoted as $\mathcal{X}_t^* = \mathcal{X}_t^j \bigcup \mathcal{B}_t^{j-1}$. The initial cluster center is obtained by utilizing the softmax output of the input target samples as follows.

\begin{equation}
    c_k^{(0)} = \frac{\sum_{x_t \in \mathcal{X}_t^*}\hat{f}_t(x_t)\hat{g}_t(x_t)}{\sum_{x_t \in \mathcal{X}_t^*}\hat{f}_t(x_t)}
\end{equation}
After computing the initial estimate of the centroids, the initial estimate of the pseudo labels $\hat{y}_t^{(0)}$ is found using the cosine distance function.
\begin{equation}
    \hat{y}_t^{(0)} = \argmin_k d(\hat{g}_t(x_t), c_k^0)
\end{equation}
where $d(\cdot,\cdot)$ is the cosine distance function. After computing the initial estimates of the pseudo labels, the cluster centers are recomputed as follows.

\begin{equation}
    c_k^{(1)} = \frac{\sum_{x_t \in \mathcal{X}_t^*}\mathbbm{1}{(\hat{y}_t=k)}\hat{g}_t(x_t)}{\sum_{x_t \in \mathcal{X}_t^*}\mathbbm{1}{(\hat{y}_t = k)}}
\end{equation}
where $\mathbbm{1}(\cdot)$ is the indicator function. The final pseudo labels are computed using the updated cluster centers.
\begin{equation}
    \hat{y}_t^{(1)} = \argmin_k d(\hat{g}_t(x_t), c_k^{(1)})
\end{equation}
where $\hat{y}_t^{(1)} \in \hat{\mathcal{Y}}_t^*$. However, computing pseudo-labels this way may lead to some noisy labels.
This effect can be more pronounced in continual DA, since each target batch contains only a partial representation of the overall target distribution because batches are composed of a small number of target samples per class. We deal with noisy pseudo-labels using sample mixup, as described next.

\subsection{Sample Mixup}

\begin{table*}[htbp]
\begin{center}
\begin{tabular}{lccccccc>{\columncolor[gray]{0.8}}c}
\toprule
Method & Target &A $\longrightarrow$ D & A $\longrightarrow$ W & D $\longrightarrow$ A & D $\longrightarrow$ W & W $\longrightarrow$ A & W $\longrightarrow$ D & Mean\\
\midrule
DANN \cite{ganin2015unsupervised} & Full & 79.7 & 82.0 & 68.2 & 96.9 & 67.4 & 99.1 & 82.2 \\
SAFN+ENT \cite{xu2019larger} & Full & 92.1 & 90.3 & 73.4 & 98.7 & 71.2 & 100.0 & 87.6 \\
ALDA \cite{alda} & Full & 94.0 & 95.6 & 72.2 & 97.7 & 72.5 & 100.0 & 88.7 \\
MDD+IA \cite{mdd} & Full & 92.1 & 90.3 & 75.3 & 98.7 & 74.9 & 99.8 & 88.8 \\
GVB-GD \cite{gvb} & Full & 95.0 & 94.8 & 73.4 & 98.7 & 73.7 & 100.0 & 89.4\\
CADA-P \cite{kurmi2019attending} & Full & 95.6 & 97.0 & 71.5 & 99.3 & 73.1 & 100.0 & 89.5 \\
HDMI \cite{hdmi} & Full & 94.4 & 94.0 & 73.7 & 98.9 & 75.9 & 99.8 & 89.5 \\
SPL \cite{wang2020unsupervised} & Full & 93.0 & 92.7 & 76.4 & 98.7 & 76.8 & 99.8 & 89.6 \\
CAN+A$^2$LP \cite{zhang2020label} & Full & 96.1 & 93.4 & 78.1 & 98.8 & 77.6 & 99.8 & 90.7 \\
SRDC \cite{tang2020unsupervised} & Full & 95.8 & 95.7 & 76.7 & 99.2 & 77.1 & 100.0 & 90.9 \\
SHOT \cite{shot} & Full & 94.0 & 90.1 & 74.7 & 98.4 & 74.3 & 99.9 & 88.6 \\
\midrule
HR-SHOT (Ours) & Full & 98.2 & 97.2 & 80.0 & 99.0 & 80.2 & 99.8 & 92.4 \\
\midrule
HR-SHOT (Ours) & Cont. & 95.8 & 90.6 & 73.8 & 96.9 & 76.7 & 99.8 & 88.9 \\
ConDA (Ours) & Cont. & 94.8 & 94.7 & 79.1 & 98.4 & 77.2 & 99.8 & 90.7 \\
\bottomrule
\end{tabular}
\end{center}
\caption{Mean accuracy on the Office-31 dataset. The ConDA experiments are performed with a continual batch size of 62 and buffer size of 124 (4 samples per class).}
\label{res:office}
\end{table*}

In the context of information maximization, since we rely on pseudo-labels that are likely to be somewhat corrupted, we employ sample and label mixup \cite{zhang2017mixup} to alleviate prediction sensitivity and achieve better generalization. 
Virtual target samples $ (\Tilde{x}_t, \Tilde{y}_t)  $ are constructed via mixup as follows.

\begin{equation}
\begin{gathered}
    \Tilde{x}_t = \lambda x_t^\alpha + (1-\lambda) x_t^\beta \\
    \Tilde{y}_t = \lambda \hat{y}_t^\alpha + (1-\lambda) \hat{y}_t^\beta
\end{gathered}
\end{equation}
where $(x_t^\alpha, \hat{y}_t^\alpha)$ and $(x_t^\beta, \hat{y}_t^\beta)$ are drawn randomly from $\{\mathcal{X}_t^*, \hat{\mathcal{Y}}_t^*\}$ and $\Tilde{x}_t \in \Tilde{\mathcal{X}}_t^*$ and $\Tilde{y}_t \in \Tilde{\mathcal{Y}}_t^*$.
Also, $\lambda \in [0,1]$ is drawn from a $Beta(\rho,\rho)$ distribution, where $\rho \in (0,\infty)$.

\subsection{Adaptation Objective Function}

For our objective function, we consider the information maximization (IM) loss from \cite{gomes2010discriminative, shisha, hu, shot} to produce individually precise predictions while maintaining a global diversity of the network outputs. The IM loss is a combination of the entropy loss $\mathcal{L}_{ent}$ and equal diversity loss $\mathcal{L}_{eqdiv}$ functions shown below. 
\begin{equation}
\begin{gathered}
    \mathcal{L}_{ent}(f_t;\mathcal{X}_t) = -\mathbb{E}_{\Tilde{x}_t \in \Tilde{\mathcal{X}}_t^*} \sum_{k=1}^{C_s} \sigma_k(f_t(\Tilde{x}_t))log(\sigma_k(f_t(\Tilde{x}_t))) \\
    \mathcal{L}_{eqdiv}(f_t;\mathcal{X}_t) = \sum_{k=1}^{C_s} q_k log  \left ( \frac{q_k}{\hat{q}_k} \right )
\end{gathered}
\end{equation}
where $\sigma_k(a) = \frac{exp(a_k)}{\sum_i exp(a_i)}$ is the softmax function. Since we maintain a class-balanced buffer, we take $q_k$ as the ideally uniform mean response, such that $q_k$ is a $C_s$ dimensional vector with all values of ${1}/{C_s}$ and $\hat{q}_k = \mathbb{E}_{\Tilde{x}_t \in \Tilde{\mathcal{X}}_t^*} [ \sigma(f_t(\Tilde{x}_t)) ] $ is the mean of the softmax output for the incoming target batch and buffer samples. The equal diversity loss $L_{eqdiv}$ attempts to make network predictions equally diverse for all classes and is calculated as the KL divergence between the ideal uniform distribution and the softmax distribution from the network outputs.
Additionally, $f_t(\Tilde{x}_t) = h_t(g_t(\Tilde{x}_t))$ is a $C_s$-dim output for each virtual target sample generated by sample and label mixup.

We further minimize $\mathcal{L}_{mixup}$, the mixup cross-entropy loss  for the generated virtual target samples, shown below.
\begin{multline}
    \mathcal{L}_{mixup}(f_t;\mathcal{X}_t) = \\ - \lambda \mathbb{E}_{\Tilde{x}_t \in \Tilde{\mathcal{X}}_t^*, \hat{y}_t^\alpha \in \hat{\mathcal{Y}}_t^*} \sum_{k=1}^{C_s} \mathbf{1}_{[k=\hat{y}_t^\alpha]} log(\sigma_k(f_t(\Tilde{x}_t))) \\ - (1-\lambda) \mathbb{E}_{\Tilde{x}_t \in \Tilde{\mathcal{X}}_t^*, \hat{y}_t^\beta \in \hat{\mathcal{Y}}_t^*} \sum_{k=1}^{C_s} \bold{1}_{[k=\hat{y}_t^\beta]} log(\sigma_k(f_t(\Tilde{x}_t)))
\end{multline}
where $\hat{y}_t^\alpha$ and $\hat{y}_t^\beta$ are the respective clustering pseudolabels for samples $x_t^\alpha$ and $x_t^\beta$ such that $\Tilde{x}_t = \lambda x_t^\alpha + (1-\lambda)x_t^\beta$. Our final objective function therefore becomes,
\begin{equation}
\mathcal{L}(g_t) =  \mathcal{L}_{ent} + \gamma_1 \mathcal{L}_{eqdiv} + \gamma_2 \mathcal{L}_{mixup}
\end{equation}
where $\gamma_1$ and $\gamma_2$  are hyper-parameters. 

\begin{table*}[htbp]
\begin{center}
\setlength\tabcolsep{0.9pt}
\resizebox{\textwidth}{!}{\begin{tabular}{lccccccccccccc>{\columncolor[gray]{0.8}}c}
\toprule
Method & Target & Ar $\rightarrow$ Cl & Ar $\rightarrow$ Pr & Ar $\rightarrow$ Rw & Cl $\rightarrow$ Ar & Cl $\rightarrow$ Pr & Cl $\rightarrow$ Rw & Pr $\rightarrow$ Ar & Pr $\rightarrow$ Cl & Pr $\rightarrow$ Rw & Rw $\rightarrow$ Ar & Rw $\rightarrow$ Cl & Rw $\rightarrow$ Pr & Mean\\
\midrule
DANN \cite{ganin2016domain} & Full & 45.6 & 59.3 & 70.1 & 47.0 & 58.5 & 60.9 & 46.1 & 43.7 & 68.5 & 63.2 & 51.8 & 76.8 & 57.6 \\
ALDA \cite{alda} & Full & 53.7 & 70.1 & 76.4 & 60.2 & 72.6 & 71.5 & 56.8 & 51.9 & 77.1 & 70.2 & 56.3 & 82.1 & 66.6 \\
SAFN \cite{xu2019larger} & Full & 54.4 & 73.3 & 77.9 & 65.2 & 71.5 & 73.2 & 63.6 & 52.6 & 78.2 & 72.3 & 58.0 & 82.1 & 68.5 \\
MDD+IA \cite{mdd} & Full & 56.2 &  77.9 & 79.2 & 64.4 & 73.1 & 74.4 & 64.2 & 54.2 & 79.9 & 71.2 & 58.1 & 83.1 & 69.5 \\
CADA-P \cite{kurmi2019attending} & Full & 56.9 & 76.4 & 80.7 & 61.3 & 75.2 & 75.2 & 63.2 & 54.5 & 80.7 & 73.9 & 61.5 & 84.1 & 70.2 \\
GVB-GD \cite{gvb} & Full & 57.0 & 74.7 & 79.8 & 64.6 & 74.1 & 74.6 & 65.2 & 55.1 & 81.0 & 74.6 & 59.7 & 84.3 & 70.4 \\
HDAN \cite{hdan} & Full & 56.8 & 75.2 & 79.8 & 65.1 & 73.9 & 75.2 & 66.3 & 56.7 & 81.8 & 75.4 & 59.7 & 84.7 & 70.9 \\
SPL \cite{wang2020unsupervised} & Full & 54.5 & 77.8 & 81.9 & 65.1 & 78.0 & 81.1 & 66.0 & 53.1 & 82.8 & 69.9 & 55.3 & 86.0 & 71.0 \\
SRDC \cite{tang2020unsupervised} & Full & 52.3 & 76.3 & 81.0 & 69.5 & 76.2 & 78.0 & 68.7 & 53.8 & 81.7 & 76.3 & 57.1 & 85.0 & 71.3 \\
HDMI \cite{hdmi} & Full & 57.8 & 76.7 & 81.9 & 67.1 & 78.8 & 78.8 & 66.6 & 55.5 & 82.4 & 73.6 & 59.7 & 84.0 &  71.9 \\
SHOT \cite{shot} & Full & 57.1 & 78.1 & 81.5 & 68.0 & 78.2 & 78.1 & 67.4 & 54.9 & 82.2 & 73.3 & 58.8 & 84.3 & 71.8 \\
\midrule
HR-SHOT (Ours) & Full & 72.1 & 84.6 & 88.4 & 83.6 & 86.7 & 87.2 & 82.6 & 73.4 & 88.5 & 85.3 & 72.3 & 90.5 & 82.8 \\
\midrule
HR-SHOT (Ours) & Cont. & 65.7 & 82.2 & 85.0 & 79.8 & 80.9 & 80.7 & 77.8 & 63.5 & 85.4 & 82.0 & 64.5 & 86.2 & 77.8 \\
ConDA (Ours) & Cont. & 64.4 & 82.2 & 86.2 & 81.3 & 82.9 & 84.0 & 81.3 & 66.6 & 86.4 & 83.5 & 66.0 & 87.1 & 79.3 \\
\bottomrule
\end{tabular}}
\end{center}
\caption{Mean accuracy on the Office-home dataset. The ConDA experiments are performed with a continual batch size of 128 and buffer size of 520 (8 samples per class).}
\label{res:officehome}
\end{table*}

\begin{table*}[htbp]
\begin{center}
\setlength\tabcolsep{1.9pt}
\begin{tabular}{lccccccccccccc>{\columncolor[gray]{0.8}}c}
\toprule
Method & Target & Plane & bycycl & bus & car & house & knife & mcycle & person & plant & sktbrd & train & truck & Per class\\
\midrule
DANN \cite{ganin2016domain} & Full & 81.9 & 77.7 & 82.8 & 44.3 & 81.2 & 29.5 & 65.2 & 28.6 & 51.9 & 54.6 & 82.8 & 7.8 & 57.6 \\
SAFN \cite{xu2019larger} & Full & 93.6 & 61.3 & 84.1 & 70.6 & 94.1 & 79.0 & 91.8 & 79.6 & 89.9 & 55.6 & 89.0 & 24.4 & 76.1 \\
ALDA \cite{alda} & Full & 93.8 & 74.1 & 82.4 & 69.4 & 90.6 & 87.2 & 89.0 & 67.6 & 93.4 & 76.1 & 87.7 & 22.2 & 77.8 \\
CAN+A$^2$LP \cite{zhang2020label} & Full & 97.5 & 86.9 & 83.1 & 74.2 & 98.0 & 97.4 & 90.5 & 80.9 & 96.9 & 96.5 & 89.0 & 60.1 & 87.6 \\
SHOT \cite{shot} & Full & 94.3 & 88.5 & 80.1 & 57.3 & 93.1 & 94.9 & 80.7 & 80.3 & 91.5 & 89.1 & 86.3 & 58.2 & 82.9 \\
\midrule
HR-SHOT(Ours) & Full & 97.0 & 89.2 & 82.8 & 65.3 & 94.9 & 97.5 & 87.2 & 82.3 & 92.0 & 93.6 & 91.9 & 64.0 & 86.4 \\
\midrule
HR-SHOT(Ours) & Cont. & 96.7 & 93.8 & 85.0 & 44.3 & 97.4 & 95.9 & 79.3 & 88.1 & 94.7 & 95.3 & 89.0 & 52.3 & 84.3 \\
ConDA (Ours) & Cont. & 97.0 & 90.4 & 80.9 & 50.0 & 95.2 & 95.7 & 80.3 & 81.9 & 94.9 & 94.2 & 91.1 & 63.9 & 84.6 \\
\bottomrule
\end{tabular}
\end{center}
\caption{Mean per class accuracy on the Visda-C dataset. The ConDA experiments are performed with a continual batch size of 192 and ConDA had buffer size of 96 (8 samples per class).}
\label{res:visda}
\end{table*}

\section{Experimental Setup}
\label{sec:experimentalsetup}
\subsection{Datasets}
We use three popular DA benchmarks for our experiments: Office, Office-Home and Visda-C. 

\textbf{Office} \cite{office} is a popular small scale dataset. The dataset has 3 domains, Amazon (A), DSLR (D), and Webcam (W) with 31 object classes of items found in an office environment in each of the domains.

\textbf{Office-home} \cite{officehome} is a medium scale dataset with  4 domains, Art (Ar), Clip Art (Cl), Product (Pr), Real-World (Rw). The dataset has 65 classes of items found in everyday office and home environments.

\textbf{Visda-C} \cite{visda} is a large scale dataset with 2 domains, Synthetic (S) and Real (R). The dataset has 12 classes. The synthetic samples are generated using 3D rendering, and the real samples are taken from MS COCO dataset \cite{lin2015microsoft}.

\subsection{Implementation Details}

In the source model, we replace the ResNet  \cite{resnet} backbone of \cite{shot} with HRNet\footnote{\url{https://github.com/HRNet/HRNet-Image-Classification/releases/download/PretrainedWeights/HRNet_W48_C_ssld_pretrained.pth}} \cite{hrnet} to obtain high resolution feature maps. The rest of the network is kept unchanged from \cite{shot}. We use a bottleneck FC layer with 256 units and a batch normalization layer, as shown in Figure \ref{fig:architecture}, followed by a final task specific FC classifier and weight normalization layer, respectively. 

We train our network with SGD optimizer with 0.9 momentum. The learning rate for the layers after the HRNet backbone is set to 10 times the learning rate of the backbone. The learning rate for the backbone is set to $\eta_0=1e^{-3}$ for all datasets except for Visda-C which has a learning rate of $\eta_0 = 1e^{-4}$. We also use a learning rate scheduler $\eta = \eta_0 \cdot (1 + 10 \cdot p)^{-0.75}$ where $p$ changes from 0 to 1 as training progresses. We empirically find that $\gamma_1=1$ and $\gamma_2=0.5$ work best for all of the datasets. The number of epochs per incoming target batch for adaptation is heuristically set to 15 for Office-31 experiments, 25 for Office-home experiments, and 3 for Visda-C experiments. Parameter $\rho$ for sample mixup is set as 1.

\begin{table*}[htbp]
\begin{center}
\begin{tabular}{lccccccc>{\columncolor[gray]{0.8}}c}
\toprule
Configuration and loss function & Target & A$\rightarrow$D & A$\rightarrow$W & D$\rightarrow$A & D$\rightarrow$W & W$\rightarrow$A & W$\rightarrow$D & Mean \\
\midrule
HR-SHOT & Full &  98.2 & 97.2 & 80.0 & 99.0 & 80.2 & 99.8 & 92.4 \\
HR-SHOT & Cont. &  95.7 & 90.6 & 73.7 & 96.9 & 76.7 & 99.8 & 88.9 \\
ConDA: Buffer + $\mathcal{L}_{mixup}$ + $\mathcal{L}_{ent}$ & Cont. & 95.0 & 93.1 & 76.7 & 97.4 & 74.9 & 99.8 & 89.5 \\
ConDA: Buffer + $\mathcal{L}_{mixup}$ + $\mathcal{L}_{ent}$ + $\mathcal{L}_{eqdiv}$ & Cont. &  94.8 & 94.7 & 79.1 & 98.4 & 77.2 & 99.8 & 90.7\\
\bottomrule
\end{tabular}
\end{center}
\caption{Performance on Office-31 dataset for various loss functions with buffer. The ablation study for the continual experiments is performed with a continual batch size of 62 and ConDA had a buffer size of 124 (4 samples per class).}
\label{tab:ablation}
\end{table*} 

\section{Results And Discussion}
\label{sec:results}

\subsection{Standard DA results}
By replacing ResNet\cite{resnet} backbone with HRNet \cite{hrnet} in the SHOT model \cite{shot}, denoted as HR-SHOT in this work, we find that the UDA performance improves significantly from our baseline method SHOT and outperforms other SOTA methods. In Office-31 dataset, as seen in Table \ref{res:office}, the performance of HR-SHOT is significantly higher than the baseline SHOT. Two of the most challenging adaptations in Office-31 are D$\rightarrow$A and W$\rightarrow$A where HR-SHOT outperforms CAN+A$^2$LP \cite{zhang2020label} by 1.81\% and 2.56\%, respectively. In Office-home dataset (Table \ref{res:officehome}), HR-SHOT outperforms the baseline SHOT with ResNet-101 backbone by a massive 11\%, with high performance gains across all domain pairs over SHOT.
In VisDA-C as shown in Table \ref{res:visda}, HR-SHOT outperforms baseline SHOT by 4.5\%. Also, $truck$ is the hardest class of the twelve classes, and HR-SHOT outperforms CAN+A$^2$LP \cite{zhang2020label} by 3.91\%. These results clearly demonstrate that utilizing an HRNet \cite{hrnet} backbone for domain adaptation can significantly improve the generalization capabilities of the overall method.

\subsection{Continual DA results}
The continual DA results for Office-31 dataset are shown in Table \ref{res:office}. In the continual setting, a buffer of size 124 with 4 slots per class and a continual batch size of 62 are chosen for Office-31. It is notable that with the HRNet backbone, continual HR-SHOT with no buffer outperforms SHOT \cite{shot}  by 0.3\%. Furthermore, ConDA outperforms the continual HR-SHOT by 1.8\%. ConDA also outperforms or matches the performance of all SOTA methods except for SRDC \cite{tang2020unsupervised}, even though ConDA has access to only a batch of the target data at a time. 

In the Office-home dataset, both continual HR-SHOT and ConDA outperform the existing standard DA methods by a large margin. ConDA outperforms HDMI \cite{hdmi} by more than 7\% on mean accuracy. It is also notable that it achieves the best performance across all the domain pairs.

In the Visda-C dataset, both HR-SHOT and ConDA perform favorably with the SOTA methods. In terms of mean per-class accuracy, ConDA outperforms most of the existing methods, including the baseline SHOT \cite{shot} by more than 1.5\%. While CAN+A$^2$LP \cite{zhang2020label} achieves the best mean per class accuracy in this dataset, ConDA does better in the challenging $truck$ category.

\subsection{Ablation Studies}
We perform ablation studies to demonstrate the impact of various parts of our model on the Office-31 dataset shown in Table \ref{tab:ablation}. For UDA, HR-SHOT outperforms SHOT by more than 3.8\%. However, performance drops by 3.5\% for continual adaptation with HR-SHOT. ConDA with buffer, sample mixup, and entropy loss improves the performance by 0.6\% over HR-SHOT. The addition of our proposed equal diversity loss to ConDA improves the overall performance by another 1.2\%. 

\begin{figure}
\centering
\begin{subfigure}[b]{0.37\textwidth}
   \includegraphics[width=1\linewidth]{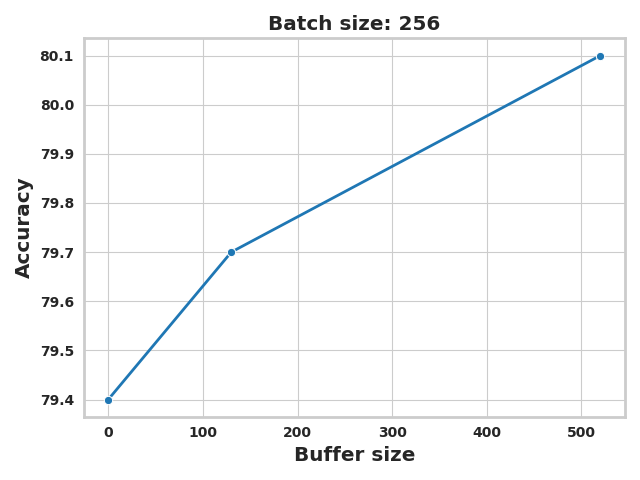}
\end{subfigure}

\begin{subfigure}[b]{0.37\textwidth}
   \includegraphics[width=1\linewidth]{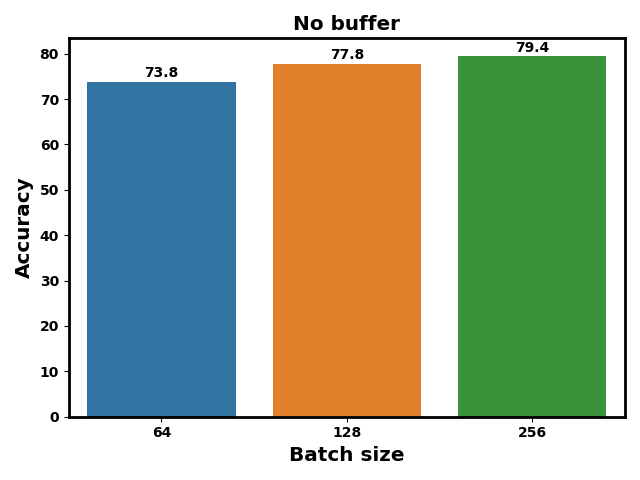}
\end{subfigure}

\caption[]{Ablation studies on Office-home dataset with varying buffer sizes (top) and varying batch sizes (bottom). 

}
\label{fig:ablation}
\end{figure}

We perform further experiments on Office-home to understand the impact of buffer sizes and batch sizes during continual adaptation as shown in Figure \ref{fig:ablation}. To study the impact of buffer size, we consider a fixed continual batch size of 256 samples and 3 different buffer sizes; no buffer, 2 samples per class, and 8 samples per class. Our findings indicate that increasing the buffer length improves performance. ConDA with a buffer size of 8 samples per class achieves 0.7\% better performance than the one with no buffer. Our study further reveals that when the number of samples in the incoming batch increases, ConDA's performance also increases. By increasing the continual batch size from 64 to 256, the overall performance improves by 5.6\%.

\section{Conclusion}
\label{sec:conclusion}

This paper introduces a new paradigm of domain adaptation where target domain data are received continually in batches for adaptation. We introduce ConDA as the first DA method to address such a setting. In ConDA, we selectively store samples in a buffer and replay them with the incoming batches to improve our network's generalization capabilities for the overall target domain. We also use sample mixup technique for data augmentation in the target domain and demonstrate its effectiveness in such a data-constrained situation. We further propose a novel loss function that improves the overall performance of our network.
We hope that this research will lay the foundation for further exploration in continual domain adaptation. 

\section*{Acknowledgements}
This research was supported in part by an AFOSR grant.
The authors acknowledge the computational resources made available by Research Computing at Rochester Institute of Technology that helped produce part of the results.

{\small
\bibliographystyle{ieee}
\bibliography{egbib}
}

\end{document}